\documentclass[runningheads]{llncs}

\usepackage[T1]{fontenc}
\usepackage{graphicx}

\usepackage{multirow}
\usepackage{booktabs}

\begin{document}
\title{Toward Robust Canine Cardiac Diagnosis: Deep Prototype Alignment Network-Based Few-Shot Segmentation in Veterinary Medicine}
%

% Anonymous
% \author{Anonymous}
% \institute{Anonymous Organization \\ 
% \email{***@********.***}}

\titlerunning{Toward Robust Canine Cardiac Diagnosis: DPANet-Based FSS in VM}

% 원래 저자명
% ----------------------------------------------------------------------------
\author{Jun-Young Oh\inst{1}\orcidID{0000-0002-4378-4082} \and
In-Gyu Lee\inst{1}\orcidID{0000-0001-7038-4618} \and \\
Tae-Eui Kam\inst{2}\orcidID{0000-0002-6677-7176} \and
Ji-Hoon Jeong\inst{1,*}\orcidID{0000-0001-6940-2700}}

\authorrunning{Jun-Young Oh et al.}

\institute{Chungbuk National University, Cheongju, 08544, Republic of Korea \\
\email{\{jy.oh,ingyu.lee,jh.jeong\}@chungbuk.ac.kr \and
Korea University, Seoul, 02841, Republic of Korea \\
\email{kamte@korea.ac.kr}}}

% -----------------------------------------------------------------------------------------------

% 짧은제목

\maketitle     % typeset the header of the contribution

\begin{abstract}
In the cutting-edge domain of medical artificial intelligence (AI), remarkable advances have been achieved in areas such as diagnosis, prediction, and therapeutic interventions. Despite these advances, the technology for image segmentation faces the significant barrier of having to produce extensively annotated datasets. To address this challenge, few-shot segmentation (FSS) has been recognized as one of the innovative solutions. Although most of the FSS research has focused on human health care, its application in veterinary medicine, particularly for pet care, remains largely limited. This study has focused on accurate segmentation of the heart and left atrial enlargement on canine chest radiographs using the proposed deep prototype alignment network (DPANet). The PANet architecture is adopted as the backbone model, and experiments are conducted using various encoders based on VGG-19, ResNet-18, and ResNet-50 to extract features. Experimental results demonstrate that the proposed DPANet achieves the highest performance. In the 2way-1shot scenario, it achieves the highest intersection over union (IoU) value of 0.6966, and in the 2way-5shot scenario, it achieves the highest IoU value of 0.797. The DPANet not only signifies a performance improvement, but also shows an improved training speed in the 2way-5shot scenario. These results highlight our model's exceptional capability as a trailblazing solution for segmenting the heart and left atrial enlargement in veterinary applications through FSS, setting a new benchmark in veterinary AI research, and demonstrating its superior potential to veterinary medicine advances.

\end{abstract}
\section{Introduction}
Recently, there has been a significant increase in the application of artificial intelligence (AI) in various fields. In particular in the medical domain, AI has made significant advances, serving as a support tool for professionals in tasks such as diagnosing patients' conditions, predicting outcomes based on medical history, and providing treatment options \cite{is1,unet,is2}. However, despite these advancements, image segmentation, one of the techniques currently used in medical AI, faces challenges due to the requirement for large amounts of labeled data \cite{doweneed}. Collecting extensive labeling data for medical images is difficult and costly due to the specialized characteristics of the task.

To address this challenge, few-shot segmentation (FSS) has emerged \cite{fss1,fss2}. The aim of FSS is to identify and segment objects using minimal labeled data. In contrast to traditional segmentation methods that rely on extensive training datasets, FSS demonstrates effective segmentation performance even in situations with limited training data. The emergence of FSS has further motivated the advancement of AI in the medical field. Prashant et al. \cite{rw1} introduced the regularized prototypical neural ordinary differential equation (R-PNODE) to perform organ image FSS. Additionally, Hao et al. \cite{rw2} proposed a self-supervised FSS network for medical images by utilizing prototypical networks \cite{pn} based few-shot learning method with the introduction of the cycle-resemblance attention (CRA) module. 

However, the majority of research on medical AI focuses primarily on human medicine, with limited application and research in veterinary medicine. In particular, myxomatous mitral valve degeneration (MMVD) occurs frequently in small, aging canines and is a disease in which heart function deteriorates due to changes in the mitral valve \cite{olddog}. This disease accounts for a significant proportion of canine mortality. One of the symptoms of MMVD, left atrial enlargement, can be visually confirmed through chest X-rays \cite{xray1}. However, the current diagnostic method, known as the vertebral left atrial score (VLAS) \cite{vlas2}, although non-invasive and relatively easy to use, is highly dependent on the experience of the examiner.

Therefore, in this paper, we propose a deep prototype alignment network (DPANet) to effectively segment the heart and symptoms of MMVD in canine chest X-ray images. Our network introduces modifications to the encoder based on PANet \cite{panet} to accurately segment the heart and left atrial enlargement areas.

Overall, the contributions of our research are as follows:
\begin{itemize}
    \item This is the first attempt to utilize FSS in veterinary medicine for segmenting the heart and left atrial enlargement part.
    \item The proposed DPANet enables effective segmentation using minimal annotated data.
    \item The proposed DPANet outperforms PANet in terms of intersection over union (IoU) and training speed.
    \item DPANet demonstrates the potential to serve as a foundational model for utilizing FSS in veterinary AI, particularly in scenarios where data are limited compared to human medical AI.
\end{itemize}

\begin{figure}[ht]
\includegraphics[width=\textwidth]{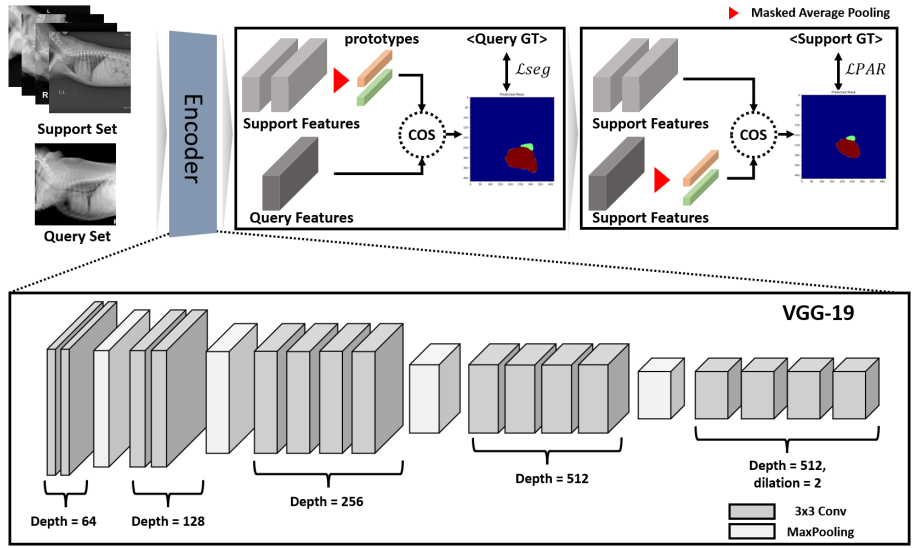}
\caption{Overview of the proposed DPANet for segmenting canine heart and left atrial enlargement part. The basic flow is similar to PANet, but we replace the encoder part with VGG-19.} \label{fig1}
\end{figure}

\section{Method}
\subsection{Overview}
PANet \cite{panet} employs VGG-16 as the encoder to extract the features required for segmentation tasks from small amounts of data. However, considering the complexity of medical images and the importance of details, we propose DPANet, which utilizes VGG-19 \cite{vgg} as a feature extractor. DPANet, with its deeper layers, aims to effectively extract intricate patterns and semantic information. DPANet operates by performing segmentation tasks based on a few data. It utilizes masked average pooling to acquire prototypes and PAR during segmentation training. The overview of DPANet is presented in Fig. \ref{fig1}.
\subsection{PAR}
PANet \cite{panet} performs FSS by introducing prototype alignment regularization (PAR). It trains class specific prototypes from a limited number of images, effectively utilizing the information between support and query instances to enhance the generalization performance of FSS. PAR extracts prototypes used to segment query images from support images and then trains these prototypes back on support images, enabling the model to learn from a small number of examples in the opposite direction. This allows the model to learn a consistent embedding space between query and support images, enhancing segmentation performance.
\subsection{DPANet}
The encoder used in PANet \cite{panet} is VGG-16 \cite{vgg}, a convolutional neural network architecture. Each convolutional layer uses a 3x3 kernel and adds non-linearity through the ReLU activation function. Additionally, pooling layers downsample image features. PANet extracts various features using VGG-16 and performs few-shot segmentation based on the extracted features. However, medical images contain complex and diverse structures, and sometimes very fine details can provide important information. Therefore, we propose an approach to better recognize and extract features from medical images by using a deeper network than VGG-16 for feature extraction. As shown in Fig. \ref{fig1}, the proposed DPANet distinguishes from PANet by using VGG-19 instead of VGG-16 as the encoder for feature extraction \cite{vgg}. When using VGG-19, DPANet adds a convolutional layer to the 3$^{rd}$, 4$^{th}$, and 5$^{th}$ blocks of the original VGG-16. As a result, a total of three convolutional layers are added to extract features. The VGG-19 empowers the model to better capture intricate patterns and semantic information. Subsequently, DPANet performs few-shot segmentation tasks using a method similar to PANet. Masked average pooling is applied to obtain prototypes from the support set. These prototypes are then used to perform segmentation on query images. The segmentation loss is defined as follows:
\begin{equation}
\mathcal{L}_{seg} =  -\frac{1}{N}\sum_{x,y}\sum_{p_{j}\in p}1[M_{q}^{(x,y)}=j]log\tilde{M}_{q;j}^{(x,y)} 
\end{equation}
$N$ represents the total number of spatial locations, $M_{q}$ denotes the ground truth segmentation mask of the query image, and $\tilde{M}_{q}$ represents the predicted segmentation mask. $P$ represents the set of prototypes. 
During segmentation training, PAR is applied to facilitate consistent embedding of prototype learning. Align loss is defined as follows:
\begin{equation}
\mathcal{L}_{PAR} =  -\frac{1}{CKN}\sum_{c,k,x,y}\sum_{p_{j}\in p}1[M_{q}^{(x,y)}=j]log\tilde{M}_{q;j}^{(x,y)} 
\end{equation}
$C$ represents the number of classes, and $K$ represents the number of support images.
The total loss for training DPANet is defined as the sum of segmentation loss and align loss. Using VGG-19 as the feature extractor, DPANet effectively captures complex patterns and semantic information, thereby strengthening the model.

\section{Experiments}
\subsection{Datasets}
To evaluate the performance of DPANet, we utilized an open dataset in AIHUB called image data(chest) for pet disease diagnosis. We utilize the experiment using 100 normal heart X-rays and 100 abnormal heart X-rays with left atrial enlargement from this dataset. All images and masks were resized to 224 x 224 pixels. During the experiments, support sets and query sets were randomly configured for each scenario and utilized for training. 

\subsection{Implementation Details}
We implemented DPANet using PyTorch. For the experiments, we used a workstation equipped with an AMD Ryzen 7 7800X3D 8-core CPU and an NVIDIA GeForce RTX 4090 GPU. Our experiments were carried out in two scenarios, 2way-1shot and 2way-5shot. The iteration was set to 2000 for all experiments. The learning rate and momentum were set to 1e-3 and 0.9. and IoU was employed as the performance evaluation metric. IoU and learning time are utilized as performance metrics.

\subsection{Results}
\begin{figure}[ht]
\centerline{\includegraphics[width=\textwidth]{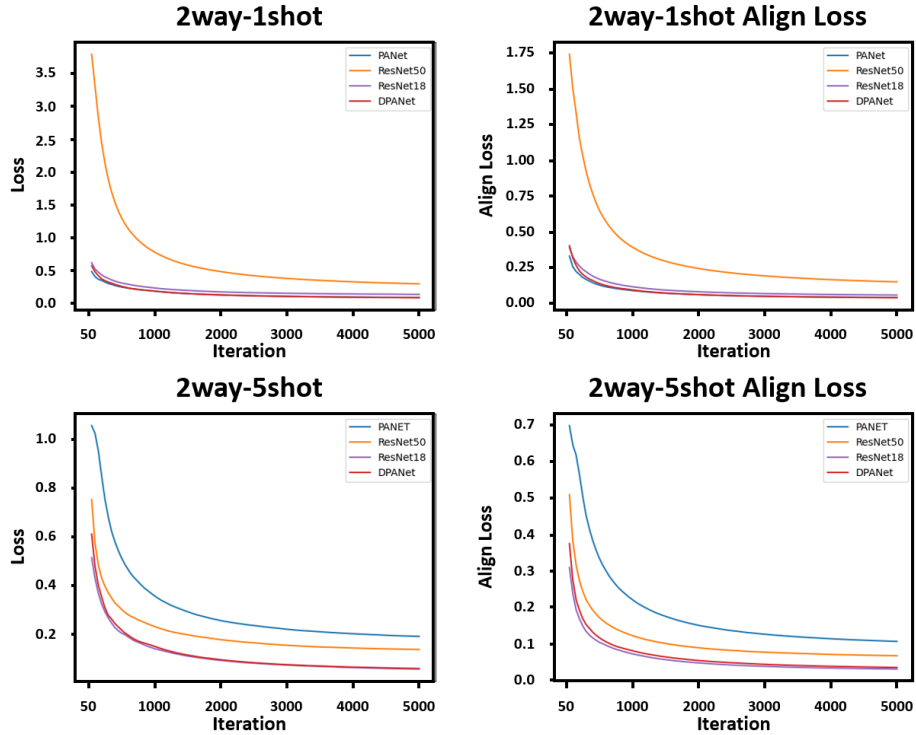}}
\caption{Loss and align loss graphs for each scenario and method. Align loss means the loss value of PAR, and loss means the training loss of DPANet combining align loss and segmentation loss.} \label{fig2}
\end{figure}

The experiment results are presented in Table \ref{tab1}, comparing our network with three methods: PANet, ResNet-18 based PANet, and ResNet-50 based PANet \cite{vgg,resnet}. All methods were trained in 5000 iterations. In the 2way-1shot scenario, our proposed DPANet achieved the highest performance with an IoU mean of 0.6966. PANet followed with an IoU mean of 0.6823, while ResNet-18 based PANet and ResNet-50 based PANet achieved performances of 0.4413 and 0.2918, respectively. In the case of learning time, PANet exhibited the fastest learning time at 6m 1s, followed by DPANet at 6m 38s, ResNet-18 based PANet at 18m 30s, and ResNet-50 based PANet at 26m 6s. In the 2way-5shot scenario, DPANet outperformed PANet in both performance and learning time. DPANet achieved the highest IoU Mean of 0.7797 and completed the training approximately 7 minutes faster than PANet, with a learning time of 27 m 59 s. PANet exhibited the second-best performance with an IoU mean of 0.7592 and a learning time of 34m 26s. ResNet-18 based PANet achieved IoU mean of 0.5302 and a learning time of 54m 56s. Finally, ResNet-50 based PANet achieved the lowest IoU mean of 0.3696 and the slowest learning time of 5h 17m 35s. Across all scenarios, DPANet consistently demonstrated the lowest values for train loss and align loss, as shown in Fig. \ref{fig2}.

\begin{table}[ht]
\caption{Results for IoU mean and learning time according to each method. The highest IoU mean and fastest learning time in each scenario are hilighted in bold.}\label{tab1}
\setlength{\tabcolsep}{10pt}
\renewcommand{\arraystretch}{1.2}
\resizebox{\textwidth}{!}{%
\begin{tabular}{l|ll|ll}
\hline
\toprule[1.5pt]
\multirow{2}{*}{Method} & \multicolumn{2}{c|}{2way-1shot}                      & \multicolumn{2}{c}{2way-5shot}                         \\ \cline{2-5} 
                        & \multicolumn{1}{l|}{IoU Mean}        & Learning Time & \multicolumn{1}{l|}{IoU Mean}        & Learning Time   \\ \midrule[1.5pt]
PANet                   & \multicolumn{1}{l|}{0.6823}          & \textbf{6m 1s} & \multicolumn{1}{l|}{0.7592}          & 34m 26s          \\ \hline
DPANet                  & \multicolumn{1}{l|}{\textbf{0.6966}} & 6m 38s         & \multicolumn{1}{l|}{\textbf{0.7797}} & \textbf{27m 59s} \\ \hline
ResNet-18 based PANet          & \multicolumn{1}{l|}{0.4413}          & 18m 39s        & \multicolumn{1}{l|}{0.5302}          & 54m 56s          \\ \hline
ResNet-50 based PANet          & \multicolumn{1}{l|}{0.2918}          & 26m 6s         & \multicolumn{1}{l|}{0.3696}          & 5h 17m 35s        \\ 
\bottomrule[1.5pt]
\end{tabular}%
}
\end{table}

\begin{figure}[ht]
\centering
\includegraphics[width=0.9\textwidth]{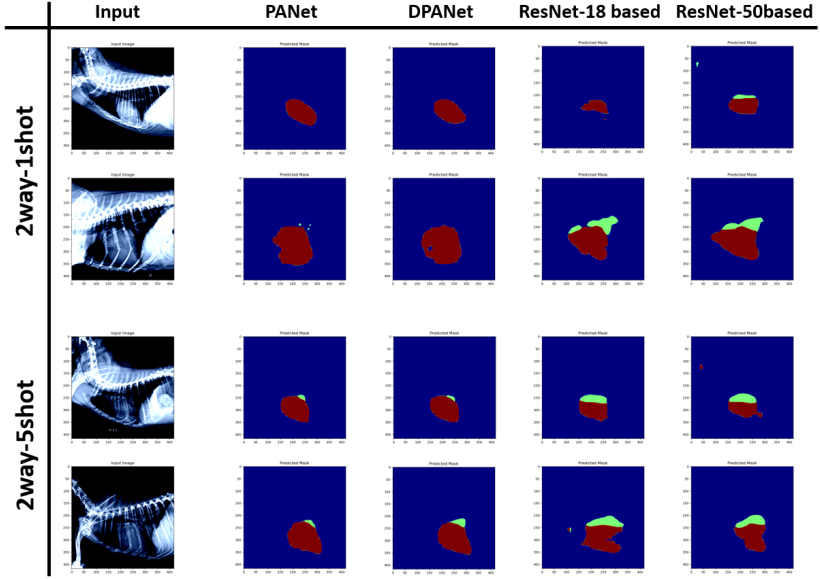}
\caption{Visualization of segmentation results of each method in 2way-1shot and 2way-5shot. Red is the part that segments the canine heart, and green is the result of segmenting the left atrial enlargement, which is one of the symptoms of cardiomegaly.} \label{fig3}
\end{figure}

Furthermore, we conducted additional qualitative experiments to further evaluate the results of segmentation, as shown in Fig. \ref{fig3}. In this figure, the red color represents the segmented areas of the heart, while the green color represents the segmented area of the left atrial enlargement. The results of the 2way-1shot segmentation demonstrated that DPANet showed accurate results of the heart shape segmentation compared to other methods. Although PANet \cite{panet} achieved precise segmentation of the heart in the first row of results, it displayed incomplete segmentation of the heart shape in the second row. ResNet \cite{resnet}-based PANet methods exhibited unstable heart shapes and incorrectly segmented areas of left atrial enlargement areas in the 2way-1shot results. The segmentation results of 2way-5shot showed superior performance for all four methods compared to 2way-1shot. In particular, the results in the fourth row indicated that the proposed DPANet outperformed other methods in segmenting areas of the left atrial enlargement area. Although ResNet-based PANet showed improved performance compared to 2way-1shot, it still exhibited unstable segmentation of the heart and left atrial enlargement areas, along with inaccurate boundaries between the two classes. These visualization results serve as additional evidence of the segmentation performance of DPANet.

Based on the results, DPANet exhibited superior performance compared to other baseline methods. In general, deeper neural networks have more layers, enabling them to extract more complex patterns and features, which is particularly useful in medical image segmentation. However, when using ResNet \cite{resnet} as an encoder, lower IoU and slower learning times were observed. These results demonstrate that it is important to select an appropriate network considering dataset size and complexity rather than unconditionally deep networks. Especially in veterinary AI where collecting large amounts of data is challenging, it is necessary to design and implement AI models taking these constraints into account.

\section{Conclusion}
We propose DPANet, a modification of the existing PANet encoder, to better recognize and extract features from medical images. Through DPANet, we effectively segment the parts of the canine heart and left atrial enlargement in a limited number of chest X-ray data. Furthermore, by performing the FSS using a canine chest X-ray for the first time, we demonstrate the potential of FSS technology as a supportive diagnostic tool for veterinarians. This can support the development of veterinary AI, which is underdeveloped compared to human medical AI. Future works include enhancing the segmentation performance of left atrial enlargement parts. After that, we will develop a canine cardiomegaly segmentation model considering various factors such as breed, gender, and weight. Moreover, we try to perform segmentation on more diverse diseases and data of pets.


\begin{thebibliography}{8}

\bibitem{is1}
Siddique, N., Paheding, S., Elkin, C. P., Devabhaktuni, V.: U-net and its variants for medical image segmentation: A review of theory and applications. IEEE Access \textbf{9}, 82031-82057 (2021)

\bibitem{unet}
Ronneberger, O., Fischer, P., Brox, T.: U-Net: convolutional networks for biomedical image segmentation. In: Navab, N., Hornegger, J., Wells, W., Frangi, A. (eds) MICCAI 2015. LNCS, vol 9351. Springer, Cham (2015). \doi{https://doi.org/10.1007/978-3-319-24574-4_28}

\bibitem{is2}
Zhou, H. Y., Guo, J., Zhang, Y., Han, X., Yu, L., Wang, L., Yu, Y.: nnformer: Volumetric medical image segmentation via a 3d transformer. IEEE Trans. Image Processing \textbf{32}, pp. 4036-4045 (2023)

\bibitem{doweneed}
Zhu, X., Vondrick, C., Fowlkes, C. C., Ramanan, D.: Do we need more training data?. International Journal of Computer Vision, \textbf{119}(1), 76-92 (2016)

\bibitem{fss1}
Liu, W., Zhang, C., Lin, G., Liu, F.: Crnet: cross-reference networks for few-shot segmentation. In: Proceedings of the IEEE/CVF Conference on Computer Vision and Pattern Recognition, pp. 4165-4173 (2020)

\bibitem{fss2}
Li, G., Jampani, V., Sevilla-Lara, L., Sun, D., Kim, J., Kim, J.: Adaptive prototype learning and allocation for few-shot segmentation. In: Proceedings of the IEEE/CVF Conference on Computer Vision and Pattern Recognition, pp. 8334-8343 (2021)

\bibitem{rw1}
Pandey, P., Chasmai, M., Sur, T., Lall, B.: Robust prototypical few-shot organ segmentation with regularized neural-ODEs. IEEE Trans. Med. Imaging \textbf{42}(9), 2490-2501 (2023)

\bibitem{rw2}
Ding, H., Sun, C., Tang, H., Cai, D., Yan, Y.: Few-shot medical image segmentation with cycle-resemblance attention. In: Proceedings of the IEEE/CVF Winter Conference on Applications of Computer Vision, p. 2488-2497 (2023)

\bibitem{olddog}
Malcolm, E. L., Visser, L. C., Phillips, K. L., Johnson, L. R.: Diagnostic value of vertebral left atrial size as determined from thoracic radiographs for assessment of left atrial size in dogs with myxomatous mitral valve disease. Journal of the American Veterinary Medical Association, \textbf{253}(8), 1038-1045 (2018)

\bibitem{xray1}
Lam, C., Gavaghan, B. J., Meyers, F. E.: Radiographic quantification of left atrial size in dogs with myxomatous mitral valve disease. Journal of Veterinary Internal Medicine, \textbf{35}(2), 747-754 (2021)

\bibitem{vlas2}
Keene, B. W., Atkins, C. E., Bonagura, J. D., Fox, P. R., Häggström, J., Fuentes, V. L., Uechi, M.: ACVIM consensus guidelines for the diagnosis and treatment of myxomatous mitral valve disease in dogs. Journal of veterinary internal medicine, \textbf{33}(3), 1127-1140 (2019)

\bibitem{panet}
Wang, K., Liew, J. H., Zou, Y., Zhou, D., Feng, J.: PANet: few-shot image semantic segmentation with prototype alignment. In: Proceedings of the IEEE/CVF International Conference on Computer Vision, pp. 9197-9206 (2019)

\bibitem{pn}
Snell, Jake, Kevin Swersky, and Richard Zemel.: Prototypical networks for few-shot learning. Advances in neural information processing systems \textbf{30} (2017).

\bibitem{vgg}
Simonyan, Karen, and Andrew Zisserman.: Very deep convolutional networks for large-scale image recognition. arXiv preprint arXiv:1409.1556 (2014)

\bibitem{resnet}
He, K., Zhang, X., Ren, S., Sun, J.: Deep residual learning for image recognition. In: Proceedings of the IEEE Conference on Computer Vision and Pattern Recognition, pp. 770-778 (2016)

%%%%%%%%%%%%%%%%%%%%%%%%%%%%%%
% \bibitem{unet3+}
% Huang, H., Lin, L., Tong, R., Hu, H., Zhang, Q., Iwamoto, Y., Wu, J.: UNet 3+: a full-scale connected UNet for medical image segmentation. IEEE International Conference on Acoustics, Speech and Signal Processing, pp. 1055-1059 (2020)

% \bibitem{fss3}
% Chen, J., Gao, B. B., Lu, Z., Xue, J. H., Wang, C., Liao, Q.: Apanet: Adaptive prototypes alignment network for few-shot semantic segmentation. IEEE Trans. on Multimedia \textbf{25}, pp. 4361-4373, (2022)

% \bibitem{xray2}
% Lee, D., Yun, T., Koo, Y., Chae, Y., Ku, D., Chang, D., Kim, H.: Change of vertebral left atrial size in dogs with preclinical myxomatous mitral valve disease prior to the onset of congestive heart failure. Journal of Veterinary Cardiology, \textbf{42}, 23-33 (2022)

% \bibitem{vlas1}
% Hansson, K., Häggström, J., Kvart, C., Lord, P.: Left atrial to aortic root indices using two‐dimensional and M‐mode echocardiography in cavalier King Charles spaniels with and without left atrial enlargement. Veterinary Radiology \& Ultrasound, \textbf{43}(6), 568-575 (2002)
%%%%%%%%%%%%%%%%%%%%%%%%%%%%%%%

\end{thebibliography}
\end{document}